\documentclass[11pt,x11names]{article}
\usepackage{setspace}
\usepackage{misc/acl2024}
\usepackage{times}
\usepackage{xcolor}
\usepackage{graphicx}
\usepackage[T1]{fontenc}
\usepackage[utf8]{inputenc}
\usepackage{microtype}
\usepackage{booktabs}
\usepackage{todonotes}
\usepackage{amsmath,amssymb}
\usepackage{dsfont}
\usepackage{cleveref}
\usepackage{xspace}
\usepackage{listings,algorithm, algorithmicx, algpseudocode}
\usepackage{soul} %
\usepackage{caption}
\usepackage{subcaption}
\usepackage{multirow,multicol}
\usepackage{ulem}
\usepackage{float}
\usepackage{array,colortbl}
\usepackage{bm}
\usepackage{xfrac}
\usepackage{tcolorbox} %
\usepackage{fontawesome} %
\usepackage[shortlabels]{enumitem}

\setlist[itemize]{noitemsep,left=0mm}

\usepackage{tabularx}

\crefname{lstlisting}{listing}{listings}
\Crefname{lstlisting}{Listing}{Listings}
\crefname{equ}{equation}{equations}
\Crefname{equ}{Equation}{Equations}
\Crefname{algorithm}{Algorithm}{Algorithms}
\crefname{example}{example}{examples}
\Crefname{example}{Example}{Examples}
\crefname{prompt}{prompt}{prompts}
\Crefname{prompt}{Prompt}{Prompts}
\DeclareCaptionType{example}[Example][List of examples]
\DeclareCaptionType{prompt}[Prompt][List of prompts]
\DeclareCaptionType{equ}[Equation][List of equations]

\usepackage{courier}
\usepackage{marginnote}

\definecolor{TodoColor}{rgb}{1,0.7,0.6}
\definecolor{TodoColor2}{HTML}{AACCAA}
\definecolor{Highlight1}{HTML}{118811}
\definecolor{Highlight2}{HTML}{A06820}

\newcommand{\xcomet}[1][]{xCOMET\textsubscript{#1}\xspace}

\newcommand{\esaai}{ESA\textsuperscript{AI}\xspace}
\newcommand{\mqmwmt}{MQM\textsuperscript{WMT}\xspace}

\newcommand{\hlc}[2][yellow]{{%
    \colorlet{foo}{#1}%
    \sethlcolor{foo}\hl{#2}}%
}

\makeatletter\def\Hy@Warning#1{}\makeatother
\let\svthefootnote\thefootnote
\newcommand\blankfootnote[1]{%
  \let\thefootnote\relax\footnotetext{#1}%
  \let\thefootnote\svthefootnote%
}

\title{AI-Assisted Human Evaluation of Machine Translation}

\newcommand{\tstar}{$^{\bigstar}$}

\author{%
    Vilém Zouhar\tstar$^1$ \qquad
    Tom Kocmi\tstar$^2$ \qquad
    Mrinmaya Sachan$^1$
    \\[1em]
    $^1$ETH Zürich \hspace{3cm} $^2$Microsoft \hspace{1cm}\\[-0.2em]
    {\tt\fontsize{9}{8}\selectfont
     \{\href{mailto:vzouhar@inf.ethz.ch}{\color{black} vzouhar},%
     \href{mailto:msachan@inf.ethz.ch}{\color{black} msachan}\}%
     @inf.ethz.ch
     \quad
     \href{mailto:tomkocmi@microsoft.com}{\color{black} tomkocmi@microsoft.com}
    }
}

\begin{document}
\maketitle
\blankfootnote{\tstar Equal contributions.}
\blankfootnote{
\,$^0$Code and collected data at:\\
\null\hfill \href{https://github.com/wmt-conference/ErrorSpanAnnotation}{github.com/wmt-conference/ErrorSpanAnnotation} 
}

\begin{abstract}
Annually, research teams spend large amounts of money to evaluate the quality of machine translation systems \citep[WMT,][inter alia]{kocmi-etal-2024-findings}.
This is expensive because it requires a lot of expert human labor.
In the recently adopted annotation protocol, Error Span Annotation (ESA), annotators mark erroneous parts of the translation and then assign a final score.
A lot of the annotator time is spent on scanning the translation for possible errors.
In our work, we help the annotators by pre-filling the error annotations with recall-oriented automatic quality estimation. 
With this AI assistance, we obtain annotations at the same quality level 
while cutting down the time per span annotation by half (71s/error span $\rightarrow$ 31s/error span).
The biggest advantage of the \esaai protocol is an accurate priming of annotators (pre-filled error spans) before they assign the final score.
This alleviates a potential automation bias, which we confirm to be low.
In our experiments, we find that the annotation budget can be further reduced by almost 25\% with filtering of examples that the AI deems to be likely to be correct.
\end{abstract}

\section{Introduction}
\label{sec:introduction}

The quality of machine translation (MT) systems is periodically evaluated by academic and industry teams to measure progress and inform product deployment decisions.
This undertaking at scale, such as the WMT campaigns \citep[][inter alia]{kocmi-etal-2023-findings,kocmi-etal-2024-findings}, is extremely expensive.
For high-quality systems, expensive high annotation quality is increasingly required to distinguish which system is truly better.
Despite recent advancements in automated metrics \citep{freitag-etal-2023-results}, the metrics remain misaligned with the ideal measure of text quality and human evaluation remains the most accurate, reliable, and ultimate standard.

\begin{figure}[t]
\centering

\includegraphics[width=\linewidth]{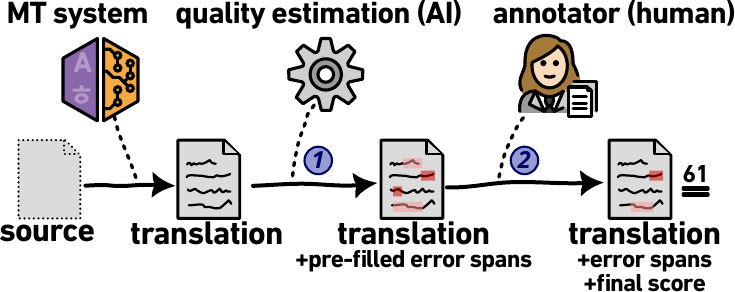}

\vspace{3mm}

\includegraphics[width=\linewidth]{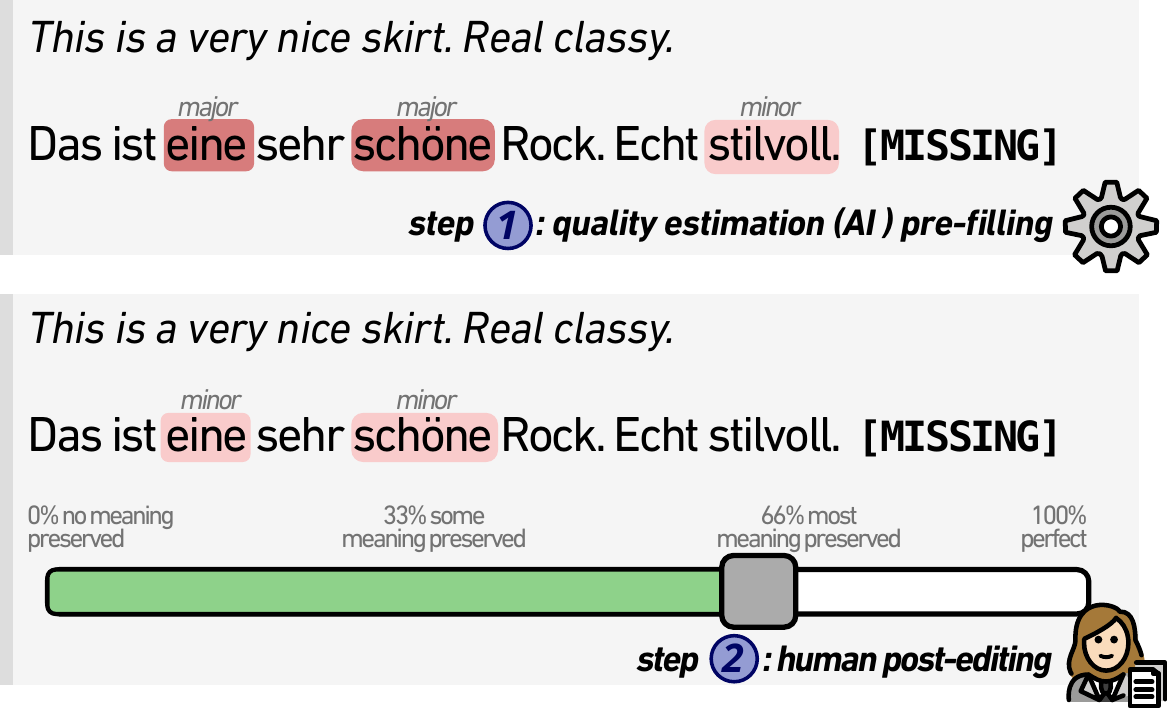}

\vspace{-3mm}

\caption{The pipeline (top) and annotation user interface (bottom) with Error Span Annotation pre-filled with AI. In the example, the user: (1) lowered the severity of the gender agreement error, (2) removed incorrectly marked error span, and (3) assigned the final score.}
\label{fig:appraise_screenshot}

\vspace{-5mm}
\end{figure}

Human evaluation protocols range from ranking different system outputs against each other \citep{novikova-etal-2018-rankme}, to assigning scores (direct assessment, DA, \citealp{graham-etal-2015-accurate}), or marking specific error spans, types, and their severities (Multidimensional Quality Metrics, MQM, \citealp{lommel2014multidimensional,freitag-etal-2021-experts}).
\citet{kocmi-etal-2024-error} simplified the MQM protocol into Error Span Annotation (ESA), which focuses on the error span severities and not the actual error types.
At the end, the annotators additionally assign a final score to the translation.
The ESA protocol thus combines the objective diagnostic qualities of MQM (error spans), with the speed and evaluation focus of DA (scoring).
One of the problems of all the existing annotation protocols is either their very high cost or low quality.
In this work, our aim is to make the MT evaluation process with ESA less expensive.

Human translation already benefits from human-AI collaboration \citep{zouhar-etal-2021-neural}.
In this work, we propose that human evaluation of MT can benefit from AI assistance in a similar way.
Despite the risk of automation bias (blind acceptance of AI suggestions), human-AI collaboration can be faster and more accurate than human or AI alone \citep{bondi2022role}.
Thus, instead of showing annotators just the source and the system translation, we pre-fill the translation with error annotations from an AI system (\Cref{fig:appraise_screenshot} bottom).
This is motivated by a lot of human labor being spent on finding possible errors, and we fill this with a recall-focused quality estimation system that produces error annotations.
The users still edit the error spans, but now spend less effort scanning the translation for possible errors.
This setup, which we call \textbf{\esaai}, is enabled by advancements in quality estimation systems \citep{guerreiro2023xcomet,fernandes-etal-2023-devil,kocmi-federmann-2023-gemba}, which provide accurate initial error spans.
The advantage of \esaai comes not only from the error span suggestions but also from priming the user with possible translation errors before assigning the final score.

To test our setup, we conduct an annotation campaign for translation evaluation with the ESA and \esaai protocols.
We compare these protocols on speed, inter-/intra-annotator agreement, quality control success, but also on a new meta-metric {subset consistency accuracy}.

\subsection*{Key findings}
The \esaai protocol yields on average 1.6 error spans per translation segment, in contrast to 0.5 for human-only ESA.
Although the overall \esaai annotation time is only slightly lower than that of ESA (58s$\rightarrow$52s/segment), \esaai halves the time per error span annotation (71s$\rightarrow$\hspace{0mm}31s/error span).
This is because the output of \esaai has more than three times the error spans per segment than in ESA.

In most of the cases where the AI did not predict any errors, the annotators did not add any new error span, confirming high recall of the AI.
We also find that we can prefilter such examples from the evaluation, save up to 24\% of the budget, and the evaluation results will be almost identical.
In addition, because of the unified priming, the annotators also become more self-consistent and have higher inter-annotator agreement, suggesting higher annotation quality.
Ultimately, this allows for a lower number of annotations required to arrive at the same system ranking (high subset consistency accuracy).

\begin{figure}[htbp]
\includegraphics[width=\linewidth]{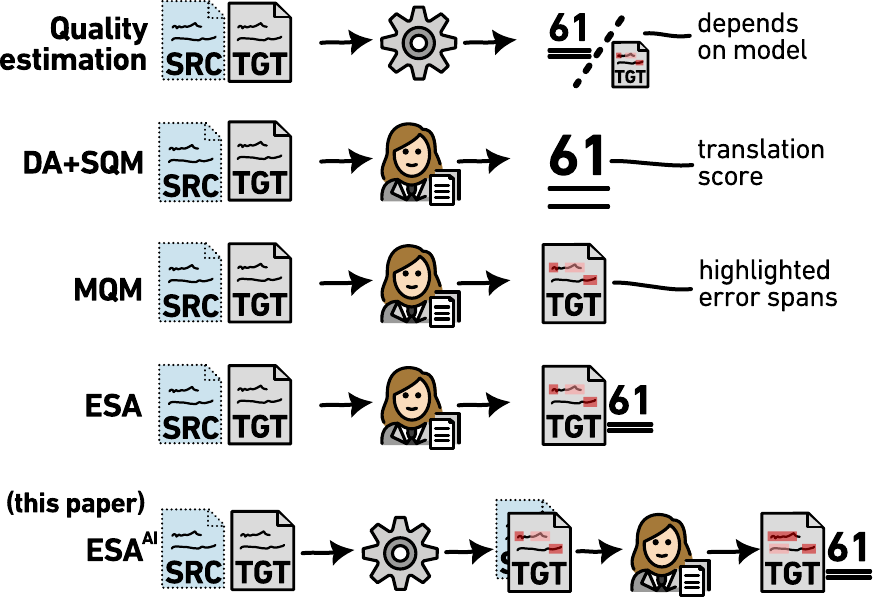}

\caption{Overview of inputs and outputs of various MT evaluation approaches. Quality estimation (QE) is automated and produces for each segment either a single score or a list of errors. DA+SQM, MQM, ESA and \esaai are human annotation protocols. \esaai (this paper) is semi-automated and happens in two-steps: quality estimation pre-annotation and human annotation.}
\label{fig:annotation_modality}
\vspace{-4mm}
\end{figure}

\section{Related Work}
\label{sec:related_work}

\paragraph{Human evaluation.}
One of the goals of MT evaluation is to compare systems to inform decisions such as which system to deploy or which machine learning method works the best.
There are two ways in which to evaluate translation quality: with automated metrics or with human labor.
Reference-based {metrics} compare the system translation to the gold human translation.
They do not always correspond to the human perception of quality and can also introduce evaluation bias \citep{freitag-etal-2020-bleu,freitag-etal-2023-results,zouhar-bojar-2024-quality}.
Reference-less approaches, known as \textbf{quality estimation (QE)}, do not have the reference bias problem, but also do not always correlate with human judgement \citep{freitag-etal-2023-results,zouhar-etal-2024-fine,falcao-etal-2024-comet-low} because the task is more difficult.
In higher-stakes settings, human annotators are thus always employed to reliably judge the translation quality.

We depict the various human evaluation protocols in \Cref{fig:annotation_modality}.
The simplest option for human evaluation is to show annotators the source and the translation and ask them to give a number from 0 to 100 \citep[DA and DA+SQM,][]{graham-etal-2015-accurate,kocmi-etal-2022-findings}.
This has the issue of low reliability and agreement.
To make annotations more objective, one can ask annotators to mark specific errors in the translation \citep[Multidimensional Quality Metrics, \textbf{MQM},][]{lommel2014multidimensional, freitag-etal-2021-experts}.
The marking is done based on their \textbf{severity} (e.g. minor or major) but also type (e.g. ``inconsistent terminology'').
This requires well-trained professional annotators and is thus expensive.
In addition, this protocol does not yield scores, but only error spans, which are turned into the final score with a hand-crafted formula that can introduce additional problems.
With some exceptions, the score computation from spans is a sum across all errors with -1 for minor and -5 for major.

\paragraph{Error Span Annotation (ESA).}
To simplify the MQM process and align it with the goal of objective translation quality assessment, \citet{kocmi-etal-2024-error} proposed \textbf{ESA}, which asks the annotators to provide only the error severity (not its type) but also a final translation score.
Because the error type is not required, the whole annotation is faster and non-experts can be employed because the annotators do not need to know the error type ontology.
This combines both DA and MQM in that the annotators are primed with their marked errors to provide high-quality final scoring.
The modalities are depicted in the penultimate row in \Cref{fig:annotation_modality}.

\paragraph{AI Assistance.}
Previous work shows that annotators can benefit from AI assistance \citep{devarajan2023ai,corals}.
However, the use of AI in evaluation is not straightforward because AI might bias the user or induce overreliance \citep{buccinca2021trust} where the annotator mindlessly accepts AI suggestions.
This can happen because annotators usually have a financial incentive to optimize their work.
In addition, \citet{veselovsky2023artificial} showed that up to 46\% of the annotators used LLMs for summarization.
Including AI assistance in the annotation directly could therefore decrease the use of undisclosed tools.
See \Cref{fig:annotation_modality} for a description of how this combines the quality estimation pipeline and the ESA protocol.

\paragraph{Quality estimation.}
Our AI assistance relies on a quality estimation (QE) system that marks error spans in the output.
Specifically, given the source and only the translation (i.e., not the reference), the QE produces error span annotations (see \Cref{fig:annotation_modality}).
Because it is not dependent on the reference, it can also be used in more setups, e.g. where the reference is only being created through this process.
Despite the history of quality estimation, such an explainable QE has become popular only recently \citep{fomicheva-etal-2021-eval4nlp}.
The most popular QE systems are
\xcomet \citep{guerreiro2023xcomet}, AutoMQM \citep{fernandes-etal-2023-devil} and GEMBA \citep{kocmi-federmann-2023-gemba}.

\textbf{GEMBA}, the QE system that we use, is based on prompting a GPT-4 model and therefore easily adaptable to new scenarios.
With a one-shot example in the target language, the model is prompted to provide a list of MQM-like errors.
We only use the error spans and severities of this model.
See the full prompts in \Cref{sec:gemba_prompts}.

The QE system is not always correct, but the output is vetted by a human annotator.
Compared to humans, the QE system is recall-focused, thus erring on the side of highlighting spans that are not erroneous.
Removing false positives is easier and faster for a human annotator than scanning the whole translation for false negatives.
The QE thus still offloads some of the work that a human would do and better primes the annotators for final score evaluation.

\section{Machine Translation evaluation with\linebreak Human-AI collaboration}

With high-quality machine translation systems, distinguishing which one is the best is increasingly difficult, requiring experts to annotate more and more samples.
Some parts of the human expert evaluation do not require full attention or can be automated.
With this, we reframe human evaluation as a computer-assisted annotation task to allow for future-proof scaling where competing systems' quality requires more evaluated samples.

We now describe the technical details needed for exact replication of the study.

\paragraph{Pipeline.}
We implement our study %
in Appraise \citep{federmann-2018-appraise} and use GEMBA, a GPT-based quality estimation system.
We adapt the Error Span Annotation (ESA) protocol (\Cref{sec:related_work}), where errors are marked on character level and annotated as either minor or major.\footnote{
\null\hfill\textbf{Minor}: style/grammar/lexical choice could be better;\\
\null\hfill\textbf{Major}: changes meaning, lowers usability. See \Cref{sec:user_guidelines}.}
The initial error markings are done by the AI and then post-edited by annotators.
Subsequently, the annotators manually assign a final score on the scale from 0\% to 100\% (not with AI) ranging from ``no meaning preserved'' to ``perfect''.
See interface screenshot in Appendix \Cref{fig:appraise_screenshot_full} and guidelines in \Cref{sec:user_guidelines}.
The error annotation part thus works as a primer for the annotators to give more accurate scores.
The complete pipeline is shown in \Cref{fig:appraise_screenshot} (top).
We run the \esaai setup twice with a different set of annotators to be able to determine the inter-annotator agreement and annotation stability.
Finally, we request about 30\% of annotators to redo their work two months later to estimate the intra-annotator agreement, also known as self-consistency.
We hire 21 annotators that are professional translators and native in the target language, German.

\paragraph{Dataset and collected data.}
We use the data of WMT23 Metrics Shared Task \citep{freitag-etal-2023-results} which has been annotated with MQM and ESA.
The \href{https://www2.statmt.org/wmt23/}{Conference on Machine Translation} annually asks research and industry teams to submit their machine translation systems.
These systems are then evaluated with human experts to determine the final system ranking.
This ranking is useful, among other things, for measuring research trends and overall improvements.
However, because the submitted systems are state-of-the-art and recently close to human quality, arriving at the ranking requires more and more annotations.
This motivates finding annotation protocols that speed up the annotations without sacrificing quality.

For maximum compatibility, we use the set-up of \citet{kocmi-etal-2024-error}.
We focus on English$\rightarrow$ German where 13 translation sets were submitted, one of which is the human reference translation and others machine translation systems.
For each set, we have 207 segments (average 18 words per segment) from 74 source documents.
Each annotator is assigned a number of segments from various sets and evaluates them with ESA or \esaai.

\section{Analyzing \esaai Efficacy}

To evaluate the new \esaai annotation pipeline, we consider two main aspects:
(1) the annotation process, including its reliability and human effort, and (2) its usefulness for machine translation systems comparison and costs.

\subsection{\esaai Evaluation Process}
\label{sec:quality_of_annotations}
\vspace{-1mm}

\paragraph{Collected data distribution.}
We first examine the high-level distribution of the data collected in \Cref{tab:overview_segment_count}.
For \esaai, the total number of annotated error spans is three times higher than for ESA, which is due to the high number of annotations suggested by the QE system.
The split between minor and major errors is similar, although \esaai annotators prefer major errors as opposed to ESA, even slightly more than those produced by the QE system.
Finally, the overall translation score is lower for \esaai than for ESA alone.
This is potentially caused by the priming effect of initially annotated error spans by the QE which highlight the negative aspects of the translation.

\begin{table}[t]
\centering
\resizebox{\linewidth}{!}{

\renewcommand{\arraystretch}{1.15}
\begin{tabular}{l>{\hspace{-3mm}}ccc}
\toprule
& \bf \#errors & \bf Minor/Major & \bf Score \\
\midrule
ESA & 0.45 & 63\%\,/\,37\% & 81.8 \\
\esaai & 1.63 & 54\%\,/\,46\% & 76.7 \\
QE {\small (automated)} & 1.51 & 55\%\,/\,45\% & $\times$ \\
\bottomrule
\end{tabular}
}
\caption{
Average number of error spans and scores across ESA, \esaai, and the QE system (automated).
Because of the pre-annotations, the output of \esaai is much more errors than ESA alone.
}
\label{tab:overview_segment_count}
\end{table}

\begin{table}[t]
\centering
\begin{tabular}{lr}
\toprule
\bf Operation & \bf Frequency \\
\midrule
Severity change & 12.0\% \\
\small \hspace{15mm} Increase severity & \small 60.0\% \\[-0.3em]
\small \hspace{15mm} Decrease severity & \small 40.0\% \\[0.65em]
Move span $\leq$5  & 13.1\% \\
Move span $\leq$10 & 17.2\% \\
Move span $\leq$20 & 23.3\% \\
\multirow{2}{*}{Resize} \small \hspace{4mm} Increase error span size & \small 21.5\% \\[-0.3em]
\small \hspace{15mm} Decrease error span size & \small 78.5\% \\
\bottomrule
\end{tabular}
\caption{Distribution of two \esaai post-editing types: changing the severity, and moving the error span. A span is considered to be \textit{moved} if the distance between old and new endpoints is at most 5, 10, or 20 characters. Many of the QE errors are only misplaced or have the wrong severity. See specific cases in \Cref{tab:post_edit_examples}.}
\label{tab:post_edit_types}
\vspace{-1mm}
\end{table}

\begin{table*}[htbp]
\newcommand{\exAA}{\textbf{Source}}
\newcommand{\exAB}{\textbf{QE}}
\newcommand{\exAC}{\textbf{\esaai}}
\centering
\renewcommand{\arraystretch}{0.95}
\resizebox{\linewidth}{!}{
\begin{tabular}{c>{\small}r>{\small}l}
\toprule
\multirow{3}{*}{\shortstack{\bf Increase\\ \bf severity}}
& \exAA & The physics are terrible and the people that created the game won't do anything about it \\
& \exAB & Die \hlc[red!20]{Physik} ist schrecklich und die Leute, die das Spiel entwickelt haben, werden nichts dagegen tun \\
& \exAC & Die \hlc[red!50]{Physik} ist schrecklich und die Leute, die das Spiel entwickelt haben, werden nichts dagegen tun \\
\cmidrule{1-2}
\multirow{3}{*}{\shortstack{\bf Decrease\\ \bf severity}}
& \exAA & Will not buy Mr. Coffee again \\
& \exAB & Ich kaufe Mr. \hlc[red!50]{Kaffee} nicht mehr. \\
& \exAC & Ich kaufe Mr. \hlc[red!20]{Kaffee} nicht mehr. \\
\cmidrule{1-2}
\multirow{3}{*}{\shortstack{\bf Move}}
& \exAA & However, I hate classes on fine arts and literature, and my school history bears it out. \\
& \exAB & Aber ich hasse \hlc[red!20]{Kunst} und Literatur, und meine Schulgeschichte bestätigt es. \\
& \exAC & Aber ich hasse Kunst und Literatur, und meine Schulgeschichte bestätigt es. \hlc[red!20]{[missing]}\\
\cmidrule{1-2}
\multirow{3}{*}{\shortstack{\bf Resize}}
& \exAA & [$\ldots$] I'm not sure if that would work for this. \\
& \exAB & [$\ldots$] ich bin mir nicht sicher, ob das für \hlc[red!20]{diesen Zwec}k funktionieren würde. \\
& \exAC & [$\ldots$] ich bin mir nicht sicher, ob das für \hlc[red!20]{diesen Zweck} funktionieren würde. \\
\bottomrule
\end{tabular}
}
\captionof{example}{Several post-editing operations from the collected data. Changing the severity (\hlc[red!20]{minor} and \hlc[red!50]{major}) is a very fast operation (only clicking the span), while moving and resizing are slow (removing the error span and creating a new one in its place takes up more of the annotator's time).
}
\label{tab:post_edit_examples}
\vspace{-2mm}
\end{table*}

\paragraph{What post-edits do annotators make?}
Not all post-editing operations are of equal value.
For example, moving the error span by a few characters to the left is less important than adding a new error span for a missing translation.
We point out two post-editing types: (1) changing the error span severity, and (2) editing the error span boundaries (\Cref{tab:post_edit_types}).
In 11\% of the cases, the users only changed the severity.
This is important from the workflow perspective because it only requires clicking on the error span.
In many cases, the error span was only moved.
Time-wise, this is more expensive because it requires the original error span to be removed and a new one created in its place.
This operation can be skipped because it does not contribute to the ESA score.
Therefore, the annotators could be instructed more specifically not to try to post-edit errors as long as they are approximately correct.
See \Cref{tab:post_edit_examples} for post-editing types.

\begin{figure}[htbp]
\includegraphics[width=\linewidth]{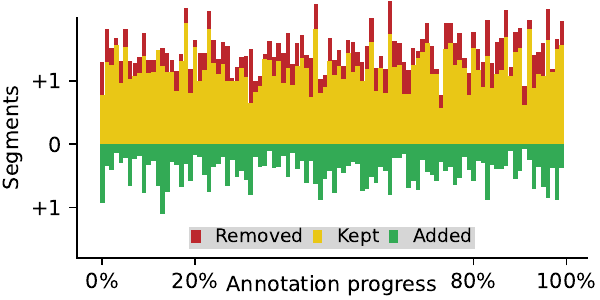}
\caption{Number of removed/kept/added error spans from the QE system with respect to annotator progress. The amount and type of work remains constant.}
\label{fig:overreliance}
\end{figure}

\paragraph{Do annotators blindly accept AI hints?}
Gradual overreliance \citep{holford2022design} is a type of habituation or automation bias that arises through repetition of non-problematic examples, such as cancer diagnosis which is dominantly negative.
Especially when there are no immediate repercussions, the annotator might be tempted to only confirm the AI suggestion without actually doing any post-editing work.
As a result, they would either confirm a span that is not an error or miss part of the translation that is not highlighted by AI but is, in fact, erroneous.
We first examine this through the perspective of changes in annotator's behavior through the annotation progression.
In \Cref{fig:overreliance} we show that the annotators make the same number of edits at the beginning as at the end of the task.
This excludes the possibility of a learned automation bias.
Next, we examine whether annotators are not already overreliant on AI from the beginning.

\paragraph{Do annotators pay attention?}
Attention checks, stimuli where we apriori know the correct annotation, are a mechanism to verify that the annotator does the expected job.
We use attention checks where the translation is intentionally malformed, but the QE system does not show an error (\Cref{ex:attention_check}).
Per each 100 segments to annotate, there are 12 attention checks in total, each with one perturbed span.
Each annotator sees both the attention check and the translation original, randomly shuffled.
This way, we can compare the annotator's score between the perturbed and non-perturbed versions.
The range for passing attention checks (on score, error count, or error highlight level) for ESA is 65\% and for \esaai 69\% (\Cref{tab:attention_checks}, similar to \citealp{kocmi-etal-2023-findings}).
This is despite \esaai being at a disadvantage because the segments, as in \Cref{ex:attention_check}, contain errors that are strictly not highlighted by the QE system.
Therefore, the pertrubed examples were even more out-of-distribution and the attention of the annotators in-distribution is likely higher.

\begin{figure}[t]
\medskip
\centering
\resizebox{\linewidth}{!}{
\begin{minipage}{1.15\linewidth}
\it
\texttt{\bf SRC}:\hspace{2.4mm} Sie haben gestern das Treffen wieder verschoben. \\
\texttt{\bf TGT}:\hspace{2.1mm} \hlc[red!40]{He} postponed the meeting again yesterday. \\
\texttt{\bf TGT$^\textbf{P}$}: \hlc[red!40]{He} postponed the meeting \ul{squirrels are never}.
\end{minipage}
}
\captionof{example}{An example of a perturbed translation \texttt{\bf TGT$^\textbf{P}$} based on the original system translation \texttt{\bf TGT}. The QE system correctly annotated the error span \hlc[red!40]{he} (correctly the pronoun is \textit{they}) but the \ul{perturbed part} is left intentionally unannotated as an attention check.}
\label{ex:attention_check}
\end{figure}

\begin{table}[ht]
\resizebox{\linewidth}{!}{
\begin{tabular}{llccc}
\toprule
& & \bf Original & \bf Perturbed & \bf OK \\
\midrule
\multirow{3}{*}{\bf ESA} & Score & 79.5 & 52.6 & 86\% \\
& Span count & 0.85 & 1.86 & 54\% \\
& Perturbation marked\hspace{-20mm} & & & 56\% \\[0.7em]
\multirow{3}{*}{\shortstack[l]{\bf \esaai}} & Score & 75.8 & 52.6 & 76\% \\
& Span count & 2.19 & 4.48 & 61\% \\
& Perturbation marked\hspace{-20mm} & & & 71\% \\
\bottomrule
\end{tabular}
}
\caption{Annotations assigned to perturbed attention check items (either scores or number of spans).
\textbf{OK} is percentage in how many cases the non-perturbed item received a higher score or had fewer error spans, and how often the pertrubed span was marked by the annotator.}
\label{tab:attention_checks}
\end{table}

\paragraph{Do AI mistakes affect annotators?}
Showing incongruent examples, where AI predictions are clearly wrong, has the potential to reduce the user's trust in AI and their subsequent collaboration \citep{dhuliawala-etal-2023-diachronic}.
In our case, such examples are the attention checks in which the AI intentionally misses the perturbed part.
To measure the effect of incongruent examples, we look at documents directly preceding and following the attention check.
In the document directly before 84\% of AI-suggested spans are accepted.
In contrast, documents directly after the attention check have only 73\% of acceptance of AI-suggested spans.
This is a slight decrease in trust, but does not render the collaboration ineffective.
It also shows that the annotators are sensitive to possible AI mistakes.

\begin{figure}[t]
\includegraphics[height=5.8cm]{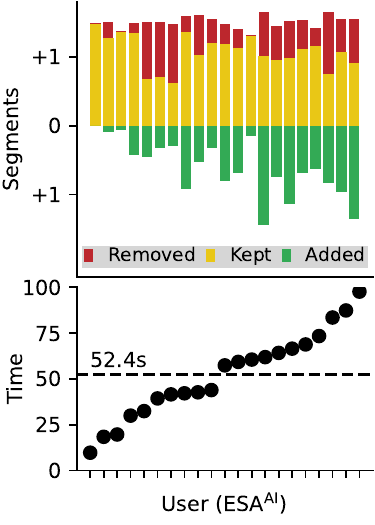}
\raisebox{1mm}{\includegraphics[height=5.8cm,trim={12mm 0 0 0},clip]{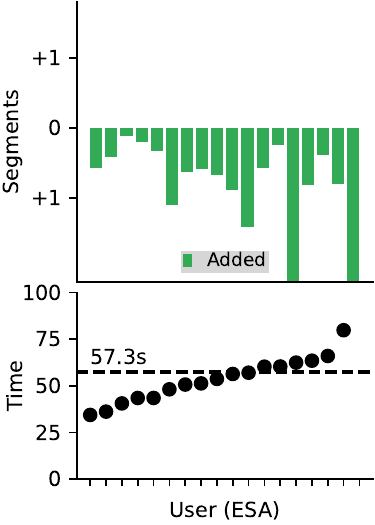}}
\caption{Annotation actions (remove/keep/add an error span) and time per segment. Each dot and bar is an annotator (sorted by time).}
\label{fig:edit_degree}
\end{figure}

\paragraph{How long do annotations take?}
One of the motivations for the AI-assisted setup is to speed up annotations and reduce costs.
The variance in individual annotator time can be explained by how much they post-edited the QE system's error span annotation (see \Cref{fig:edit_degree}).
In this aspect, \esaai has more variance than ESA, but can also be more controlled and constrained by instructing the annotators what the expected post-editing level is.
Per segment, \esaai annotators required 52s while ESA required 58s.
In addition, the time is 71s per single error span for ESA but 31s per single error span for \esaai, making the latter more efficient in detailed annotation.

\paragraph{Do annotators agree?}
For a robust and objective annotation protocol, the scores by two independent annotators should be similar and not subjective.
To test this, we ran the annotations again with different annotators.
\Cref{tab:interintra_annotator_agreement} shows that \esaai has a much larger agreement between the annotators.
For the MQM-like score computation from spans, this is due to the bias by the pre-filled error spans.
Still, the agreement is much higher also for the direct scoring, likely due to the unified priming of the annotators.
This is consistent with much higher \esaai \textit{intra}-annotator agreement, i.e. how much annotators agree with themselves.

\begin{table}[t]
\centering
\resizebox{\linewidth}{!}{
\begin{tabular}{lcccc}
\toprule
& \multicolumn{2}{c}{\bf inter-annotator}
& \multicolumn{2}{c}{\bf intra-annotator}\\
\textbf{Scoring}
& \bf ESA & \bf \esaai
& \bf ESA & \bf \esaai \\
\midrule
direct score & 0.376 & 0.533 & 0.222 & 0.486 \\
from spans & 0.327 & 0.671 & 0.282 & 0.689 \\
\bottomrule
\end{tabular}
}
\caption{Inter-annotator and intra-annotator agreement with direct scores and scores computed from error spans with MQM formula, as measured with Spearman correlation.
\esaai from spans have the highest inter-annotator agreement, which is however caused by the the QE system's pre-filling.
Still, the scores from \esaai, solely by humans, have the highest inter-annotator and intra-annotator agreement. See visualization in Appendix \Cref{{fig:intrainter_annotator_agreement}}.}
\label{tab:interintra_annotator_agreement}
\end{table}

\begin{figure}[ht]
\includegraphics[width=\linewidth]{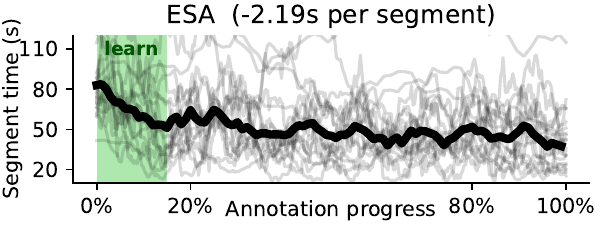}
\vspace{-2mm}

\includegraphics[width=\linewidth]{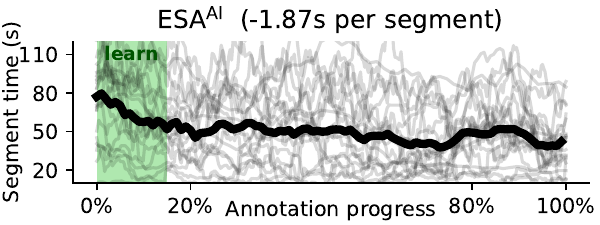}
\caption{Time per segment with respect to progression in the annontation. Each annotator is the gray faint line and their average is in black. The lines are smoothed with a window of size 15 segments. We also compute the average speed at the beginning and at the end, which yields the \textit{learned speedup}. This is how much the annotator speeds up per working on one segment.}
\label{fig:document_speedup}
\end{figure}

\paragraph{Do annotators become faster?}
With most annotations tasks, the annotators \textit{learn} to be faster.
Although the speed-up occurs throughout the entire annotation, it is mostly present in the first 15\% of segments (green box in \Cref{fig:document_speedup}).
The \esaai annotators get 1.87s faster with every segment, which is comparable to ESA.
This effect is present despite the ESA annotators being at an advantage because there were more \esaai annotators, and thus each \esaai annotator individually processed fewer segments, having less time to learn.
In addition, users in the post-editing task are more consistent.
For ESA, the user's absolute deviation from their personal average is 43.3s, while for \esaai this is only 32.1s.
This makes the human effort more consistent and predictable but also shows that the nature of the annotation task changes.

\paragraph{Why do some segments take longer than others?}
Being able to predict the expected annotation time for a segment can lead to a more efficient distribution and planning, for example, when selecting which segments to annotate at all.
We examine the correlations between the features and the segment-level time in \Cref{tab:time_linear_correlations}.
The number of words in the translation, together with the number of error spans, is a strong predictor of annotation time.
For MQM this is the highest, which can be explained by each error span requiring the most work in the MQM annotation scheme because the annotators have to also assign the error type.
The longer the whole document (number of surrounding translation segments), the lower the annotation time, which is likely due to shared context, so the annotator does not have to switch between domains and contexts.
In contrast, the \esaai annotators are slightly less affected by the translation length in contrast to ESA because the error spans are pre-highlighted.

\begin{table}[t]
\centering
\resizebox{\linewidth}{!}{
\begin{tabular}{lrrr}
\toprule
& \bf MQM & \bf ESA & \bf \esaai \\
\midrule
Progress &
 -0.12 &  -0.13  & -0.13  \\
Translation word count &
  0.30 &  0.19  & 0.16  \\
QE error spans &
  0.12 &  0.07  & 0.12  \\
Error spans &
  0.06 &  0.04  & 0.12  \\
Score &
 -0.07 &  -0.03  & -0.08  \\
Document size &
 -0.14 &  -0.17  & -0.17  \\
\bottomrule
\end{tabular}
}
\caption{Individual Pearson correlation between features and annotation times. The higher the absolute value, the more it affects the annotation time.}
\label{tab:time_linear_correlations}
\end{table}

\subsection{\esaai for Evaluation of WMT Systems}
\label{sec:wmt_clustering}

Our goal is for \esaai to be as reliable as or more reliable than ESA in ranking MT systems.
We consider \mqmwmt collected by \citet{freitag-etal-2023-results} as the human gold standard and show the system-level correlations with our protocol in Appendix \Cref{fig:clusters}.
Both ESA and \esaai have similar correlations with \mqmwmt, justifying our setup.
In \Cref{tab:cross_protocol_correlation} we show that this protocol does not stray far from existing ones in terms of segment-level rating.
Many of these cross-protocol correlations are on part with inter-annotator agreement, which is naturally the upper bound.
In particular, \esaai has a higher correlation than ESA or MQM by \citet{kocmi-etal-2024-error} alone.

\begin{table}[t]
\centering
\resizebox{\linewidth}{!}{
\begin{tabular}{lcccc}
\toprule
& \bf ESA & \bf \esaai & \bf MQM & \bf QE \\
\bf \mqmwmt  & 0.240 & 0.292 & 0.239 & 0.416 \\
\bottomrule
\end{tabular}
}
\caption{Kendall $\tau_c$ segment-level correlations between evaluation protocols. ESA and \esaai use direct scores.}
\label{tab:cross_protocol_correlation}
\end{table}

\paragraph{Can cost be further lowered?}
\label{sec:quality_of_postediting}
The goal is to speed up the annotation process without sacrificing quality.
This can be achieved by removing, or automating, redundant decisions and actions on the annotator's side.
In this segment, we do so by skipping high-quality translations for which we can predict that the annotator's would not mark any error spans.

\begin{figure}[htbp]
\centering
\includegraphics[height=3.9cm]{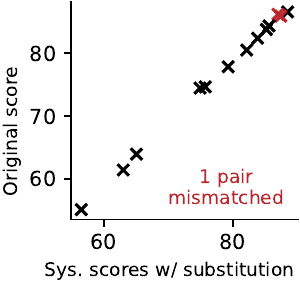}
\includegraphics[height=3.9cm,trim={10mm 0 0 0},clip]{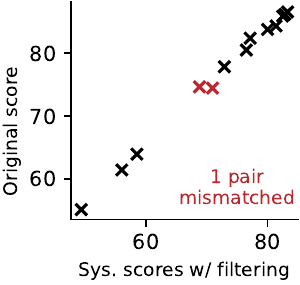}
\caption{Average system scores with either substitution or filtering of segments with no QE errors. Each cross is a single system. The two red pairs of crosses are the single pairs of systems whose ordering changes when substitution or filtering is applied.}
\label{fig:quality_gesa_prefiltering}
\end{figure}

Our QE system, GEMBA, is recall-focused, and therefore the occurrence of ``false positive'' error spans is low.
In 89\% of the cases, the QE marked the spans as having 0 errors and retained 0 errors after the annotation (first row in \Cref{tab:post_edit_distribution}), and these segments have an average score of 95.
This makes it possible to also use the QE as a prefiltering step.
If we replace all such segments with 100 (not to overfit), all but one system comparisons remain the same  (\Cref{fig:quality_gesa_prefiltering}, left).
Alternatively, one can also exclude segments for which the QE marks 0 errors for most systems, which has the advantage that we do not alter the data.
For this method, again all but one system comparison would be the same (\Cref{fig:quality_gesa_prefiltering}, right).
Pre-filtering can thus result in almost 25\% budget saving (\textasciitilde52 segments per system).

\begin{table}[htbp]
\centering
\resizebox{\linewidth}{!}{
\renewcommand{\arraystretch}{1.3}
\begin{tabular}{
>{\small}l
>{\hspace{-1mm}}c>{\hspace{-1mm}}c>{\hspace{-1mm}}c
>{\hspace{-3mm}\columncolor{gray!30}}c<{\hspace{-3mm}}
>{\hspace{-1mm}}c>{\hspace{-1mm}}c>{\hspace{-1mm}}c>{\hspace{-1mm}}c
}
\toprule
\multicolumn{1}{c}{\bf QE} &
\multicolumn{3}{c}{\bf ------Removed------} & 
\cellcolor{white} \bf \hspace{-1mm}No edit\hspace{-1mm}
& \multicolumn{4}{c}{\bf ------Added------} \\[-0.3em]
\small \#err. (freq.) &
\small $=$2 & \small $=$1 & \small $=$0 &
\cellcolor{white}
&
\small $=$0 & \small $=$1 & \small $=$2 & \small $\geq$3
\\[-0.3em]
\midrule
0 (23.8\%) & 0\% & 0\% & 100\% & 88\% & 88\% & 8\% & 2\% & 2\%\\
1 (38.0\%) & 0\% & 28\% & 72\% & 62\% & 81\% & 14\% & 3\% & 3\%\\
2 (18.8\%) & 15\% & 16\% & 69\% & 54\% & 71\% & 13\% & 9\% & 7\%\\
3 (10.4\%) & 11\% & 20\% & 62\% & 51\% & 68\% & 16\% & 7\% & 10\%\\
4 (8.9\%) & 11\% & 13\% & 69\% & 54\% & 65\% & 13\% & 10\% & 12\%\\
\bottomrule
\end{tabular}
}
\caption{Distribution of error span post-editing based on original QE-reported error spans (2nd column). Percentages in the table are proportions within the number of the QE error spans. For example, second row shows that 62\% of segments with exactly one QE error span received no post-editing from annotators and in 28\% the annotators removed the single error. ESA is comparable to \esaai.}
\label{tab:post_edit_distribution}
\smallskip
\end{table}

\paragraph{How many annotations are needed?}
\label{sec:subset_consistency}

Comparing the quality of two annotation protocols is not straightforward because of the absence of a gold standard.
We now take a practical perspective where we desire a protocol that finds the true system ordering with as few examples as possible.
This does not require any gold standard to compare with.
Instead, this approach compares a subset of the annotations of a protocol with the full set of annotations from the same protocol.

\begin{figure}[htbp]
\centering
\includegraphics[height=4.8cm]{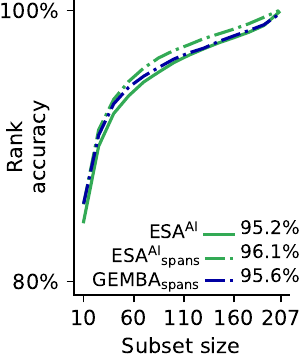}
\includegraphics[height=4.8cm,trim={10mm 0 0 0},clip]{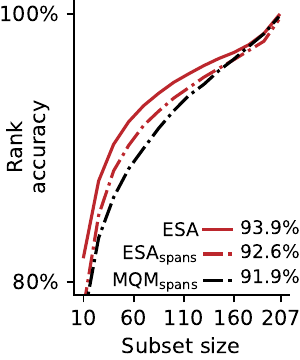}
\caption{Subset consistency accuracy of a system ranking only on a subset against ranking on full data. Percentages are averages across visible subset sizes, which corresponds to the area under the curve. See figure in tabular form in Appendix \Cref{tab:subset_consistency}.}
\label{fig:subset_consistency}
\end{figure}

With a sufficiently large evaluation, even noisy annotation schemes yield the true system ordering.
Conversely, only robust annotation schemes yield this ordering on a small scale.
We formalize this in \Cref{sec:subset_consistency_justification} to show that \esaai leads to better annotations than ESA or MQM.
We measure the accuracy of the ordering ($m_1{>_I}m_2$) of systems ($\mathcal{M}$) computed on a subset of segments ($I$) against the ordering given by the full data ($a_{m_1}{>}a_{m_2}$):
\begin{align}
\hspace{-1mm}
\textsc{Acc}(I) \overset{\text{def}}{=}
\hspace{-3mm} \sum_{m_1, m_2 \in \mathcal{M}} \hspace{-3mm}
\frac{\mathds{1}[(m_1{>_I}m_2) \Leftrightarrow (a_{m_1} {>} a_{m_2})]}{|\mathcal{M}|^2}
\nonumber \\[-1.5em]
\end{align} 
\noindent
This simulates the setup where we wish to save costs with fewer number of annotations but care about arriving at the true system ordering.
We refer to this metric as \textbf{subset consistency accuracy}, similar to split-half reliability, and in \Cref{sec:subset_consistency_justification} justify that annotation protocols with lower noise reach higher accuracies.
For each subset size, e.g. 30 source sentences, we select 1000 random subsets and compute the system ranking accuracy.

The results in \Cref{fig:subset_consistency} suggest that
the QE system, GEMBA, alone very consistent.
However, this is solely because it can be perceived as a single annotator and thus there is no inter-annotator confusion.
Overall and among human evaluation protocols, \esaai with scores has the highest stability and quality of scores.
This is not the result of automation bias because only the annotators themselves assign the final score.
In practice, this would mean that one can annotate fewer examples (e.g. 2000 for \esaai) to obtain the same system-level accuracy as lower-quality protocol (e.g. 2500 for ESA), thus further lowering the costs.

\section{Conclusion}

Our AI-assisted protocol of human evaluation of MT is \hlc[gray!0]{faster and cheaper}.
This protocol is more robust and self-consistent and increases inter-annotator agreement by priming the annotators with pre-annotated error spans.
Our analysis also shows that the annotators did not overrely on the AI and were able to \hlc[gray!0]{maintain evaluation quality}.
The inclusion of AI in evaluation also opens many options for further evaluation economy by reducing the test set size requirements.
To this end, we introduce \hlc[gray!0]{subset consistency accuracy}, which quantifies how many annotations could be saved while arriving at a similar final system ordering.

\section*{Limitations}

Despite the advantages in lower costs per error span of the presented setup, we urge practitioners not to use this approach when metrics evaluation is one of the expected tasks due to the particular bias to the used metric in the setup.
The intended application of this pipeline is purely a more efficient evaluation of the quality of the machine translation system.

Both \esaai and GEMBA rank GPT-4-5shot as the best system, a system that uses the same LLM to translate sentences as we use to generate for GEMBA.
This indicated a weakness that our approach is biased towards systems built on top of the same underlying LLM. 
\citet{liu-etal-2023-g} described this phenomenon when the same system used to generate output should not be used to also evaluate them.
This issue could be mitigated by using two different LLMs to generate error spans.

Lastly, for QE we use GEMBA, a GPT4-based system, for the quality estimation and work with WMT 2023 data.
Unfortunately, we cannot exclude the possibility of the QE system being trained on this data, though the texts and scores are kept in two separate large files with non-linear mappings.

\section*{Ethics Statement}
The annotators were paid a standard commercial translator wage in the respective country.
No personal data was collected and the data shown to the annotators was screened for potentially disturbing content.

\bibliography{misc/anthology.min.bib,misc/bibliography.bib}

\clearpage

\appendix
\begin{table}[htbp]
\resizebox{\linewidth}{!}{
\begin{tabular}{lcccc}
\toprule
\multirow{2}{*}{\shortstack{\bf Protocol/\\\bf method\,\,\,}} & \multicolumn{4}{c}{\bf Subset size} \\
 & \bf 10 & \bf 40 & \bf 115 & \bf 190 \\
\midrule
ESA$^\mathrm{AI}$ & 84.41\% & 92.38\% & 96.69\% & 98.88\% \\
ESA$^\mathrm{AI}_\mathrm{spans}$ & 85.69\% & 93.43\% & 97.46\% & 99.49\% \\
GEMBA$_\mathrm{spans}$ & 85.73\% & 93.10\% & 96.86\% & 98.94\% \\
ESA & 81.86\% & 90.26\% & 95.52\% & 98.52\% \\
ESA$_\mathrm{spans}$ & 78.11\% & 88.28\% & 94.48\% & 97.94\% \\
MQM$_\mathrm{spans}$ & 77.19\% & 86.30\% & 93.89\% & 98.50\% \\
\bottomrule
\end{tabular}
}
\caption{Specific values of \Cref{fig:subset_consistency}. Subset accuracy across annotation schemes. \esaai\textsubscript{spans} has the highest subset consistency, though this is likely biased by the spans from GEMBA, which as 100\% inter-annotator agreement. However, \esaai (direct scores) is based solely on human scorings, which has the second-highest subset consistency of any protocol.}
\label{tab:subset_consistency}
\end{table}

\section{Subset Consistency Formalization}
\label{sec:subset_consistency_justification}

This section justifies the setup in \Cref{sec:subset_consistency} and is reminiscent of the work of \citet{riley-etal-2024-finding} or split-half reliability.
A key distinction is that we are considering ranking stability with respect to the protocol itself.
We do so by bootstrapping subsets of the data.

Our goal is tho show that a protocol with lower annotation error has higher system-level ranking accuracy.
We assume that the annotation schemes are not biased towards a particular system but are noisy.
We also assume a simplified model of system performance, where the annotation output $y_{m,i}$ of system $m$ on segment $i$ can be approximated by the system ability $a_m$ (e.g. average across a real life distribution) from which segment-specific variance $d_i$ is subtracted and error term $\epsilon$ is added.
The annotation output $y_{m,i}$ is dependent on the specific annotation scheme, which is not indicated for brevity.
We would like to find the system abilities $a_m$ but we only have access to $y_{m,i}$.
This notation can also be extended to a collection of segments $I$:
\begin{align}
y_{m,i} &= a_m - d_i + \epsilon_{m,i} \\
Y_{m,I} &= \frac{\sum_{i\in I} y_{m,i}}{|I|} \\
&= a_m - \frac{\sum_{i\in I} d_i}{|I|} + \frac{\sum_{i\in I} \epsilon_{m,i}}{|I|}
\end{align}
On a large enough set of segments with the law of large numbers, we can assume $\frac{\sum_{i \in I} \epsilon_{m,i}}{|I|} \approx 0$ as $\epsilon$ is unbiased.
If we want to estimate $\epsilon_{m,i}$, we could subtract from sample $i$ the average from all dataset, $Y_{m, D}$.
Unfortunately, this would still leave the segment-specific difference $d_i$:
\begin{align}
y_{m,i} - Y_{m,D} &= - d_i + \epsilon_{m,i}
\end{align}
To separate $\epsilon_{m,i}$, we could consider subsets $I \subsetneq D$ for which $\frac{\sum_{i\in I}d_{m,i}}{|I|} \approx 0$ but $\frac{\sum_{i\in I}\epsilon_{m,i}}{|I|} \not\approx 0$.
Apart from the difficulty of finding such subsets, our goal is to have a good estimation of the ranking of the systems.
For this, we define system ordering $>_I$ given by the observed subset $I$:
\begin{align}
&m_1 >_I m_2 \nonumber\\
&\overset{\text{def}}{\Leftrightarrow}
\frac{\sum_{i\in I} y_{i, m_1}}{|I|} > \frac{\sum_{i\in I} y_{i, m_2}}{|I|} \\
&\Leftrightarrow \sum_{i\in I} y_{i, m_1} > \sum_{i\in I} y_{i, m_2} \\
&\Leftrightarrow a_{m_1}-\sum_{i \in I} d_i + \sum_{i \in I} \epsilon_{i, m_1} > \nonumber\\
&\hspace{23mm} a_{m_2}-\sum_{i \in I} d_i + \sum_{i \in I}\epsilon_{i, m_2} \\
&\Leftrightarrow a_{m_1}{+}\sum_{i \in I}\epsilon_{i, m_1} >
a_{m_2}{+}\sum_{i \in I}\epsilon_{i, m_2}
\end{align}
Notice that $>_I$ is independent of the segment-specific term $d_i$ because both systems are evaluated on the same segments.
We compare this empirical ordering with that of the true system ranking.
This is done across a set of systems $\mathcal{M}$ using pairwise accuracy, i.e. how many system pairs are ranked in the same way as by the true system ranking:
\begin{align}
\hspace{-1mm}
\textsc{Acc}(I) \overset{\text{def}}{=}
\hspace{-3mm} \sum_{m_1, m_2 \in \mathcal{M}} \hspace{-3mm}
\frac{\mathds{1}[(m_1{>_I}m_2) \Leftrightarrow (a_{m_1} {>} a_{m_2})]}{|\mathcal{M}|^2}
\nonumber \\[-1.5em] \label{eq:rank_accuracy}
\end{align} 
With higher accuracy we can assume that the relative $\epsilon$ is lower, at least for the purposes of ordering.
This is because \textcolor{Highlight1}{\bf if} the accumulated error terms are low (\ref{eq:low_error}), the indicator in \Cref{eq:rank_accuracy} is true (\ref{eq:indicator_inside}), which is \textcolor{Highlight2}{\bf equivalent} to high accuracy (\ref{eq:high_acc}):
\begin{align}
& \sum_{i \in I}\epsilon_{i, m_1} \rightarrow 0 \wedge \sum_{i \in I}\epsilon_{i, m_2} \rightarrow 0 \quad {\color{Highlight1}\bm{\Rightarrow}} \label{eq:low_error} \\
& \big( a_{m_1}{+}\sum_{i \in I}\epsilon_{i, m_1} {>}
a_{m_2}{+}\sum_{i \in I}\epsilon_{i, m_2} {\color{black}\Leftrightarrow\,} a_{m_1}{>}a_{m_2} \big) \nonumber \\[-1.2em] \label{eq:indicator_inside} \\
& {\color{Highlight2}\bm{\Leftrightarrow}} \quad \textsc{Acc}(I) \rightarrow 1 \label{eq:high_acc}
\end{align}

To obtain \textsc{Acc}, we would need to know if $a_{m_1} {>} a_{m_2}$.
In our setup, we do not know this true ranking and obtaining it would require large-scale super-human annotations.
However, for large-enough $I$, we can assume that $\frac{\sum_{i \in I} \epsilon_{m,i}}{|I|} \approx 0$.
Therefore, for the true ordering, we use the ordering by that particular annotation scheme on all data.
Now we established a link between accumulated annotation noise, $\sum_{i \in I} \epsilon_{i, m}$, and accuracy, which we can measure.

The accuracy will be high if the error terms are low and therefore the annotations are of high quality.
This can be used to measure the annotation protocol usefulness.
In addition, this has practical implications as we could solicit fewer annotations to obtain the same results as if we had more.

\section{User Guidelines}
\label{sec:user_guidelines}

The following are are annotation guidelines for our two local \esaai campaigns, which is closely based on the setup of \citet{kocmi-etal-2024-error}.

{
\fontsize{10}{11}\selectfont

\paragraph{Highlighting errors:}
Highlight the text fragment where you have identified a translation error (drag or click start \& end).
Click repeatedly on the highlighted fragment to increase its severity level or to remove the selection.
\begin{itemize}[topsep=0mm]
\item        \textbf{Minor Severity:} Style/grammar/lexical choice could be better/more natural.
\item        \textbf{Major Severity:} Seriously changed meaning, difficult to read, decreases usability.
\end{itemize}
If something is missing from the text, mark it as an error on the \texttt{\bf [MISSING]} word.
The highlights do not have to have character-level precision. It's sufficient if you highlight the word or rough area where the error appears.
Each error should have a separate highlight.
        
\paragraph{Score:} After highlighting all errors, please set the overall segment translation scores. The quality levels associated with numerical scores on the slider:
\begin{itemize}[topsep=0mm]
\item 0\%: No meaning preserved: Nearly all information is lost in the translation.
\item 33\%: Some meaning preserved: Some of the meaning is preserved but significant parts are missing. The narrative is hard to follow due to errors. Grammar may be poor.
\item 66\%: Most meaning preserved and few grammar mistakes: The translation retains most of the meaning. It may have some grammar mistakes or minor inconsistencies.
\item 100\%: Perfect meaning and grammar: The meaning and grammar of the translation is completely consistent with the source.
\end{itemize}

\newpage

\begin{figure*}[htbp]
\includegraphics[width=\linewidth]{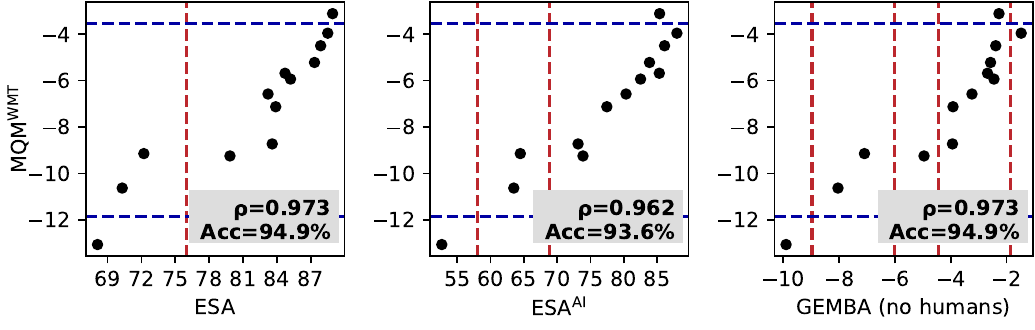}
\caption{Each point is a system, with original \mqmwmt scores on the $y$-axis against ESA, \esaai, and GEMBA before post-editing.
Stripped lines indicate cluster separations with alpha threshold 0.05. 
Numbers show Spearman's correlations between the specific protocol and \mqmwmt.
ESA and \esaai have comparable system-level accuracy and correlations with \mqmwmt, making them equal in quality in this aspect.
}
\label{fig:clusters}
\end{figure*}

\begin{figure*}[htbp]
\includegraphics[width=\linewidth]{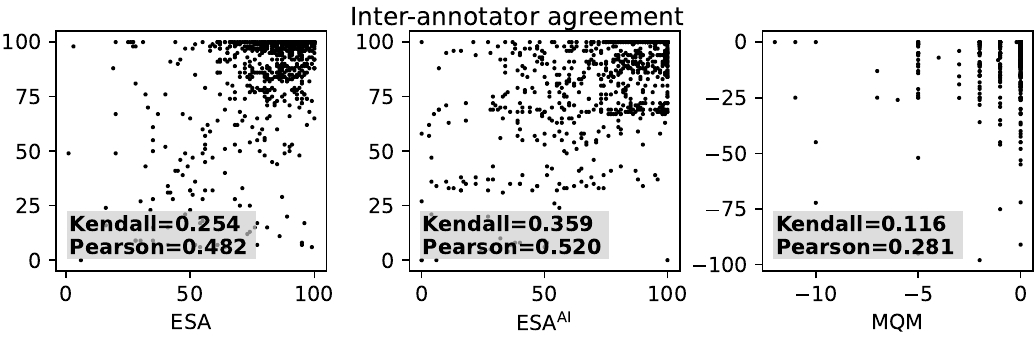}
\includegraphics[width=\linewidth]{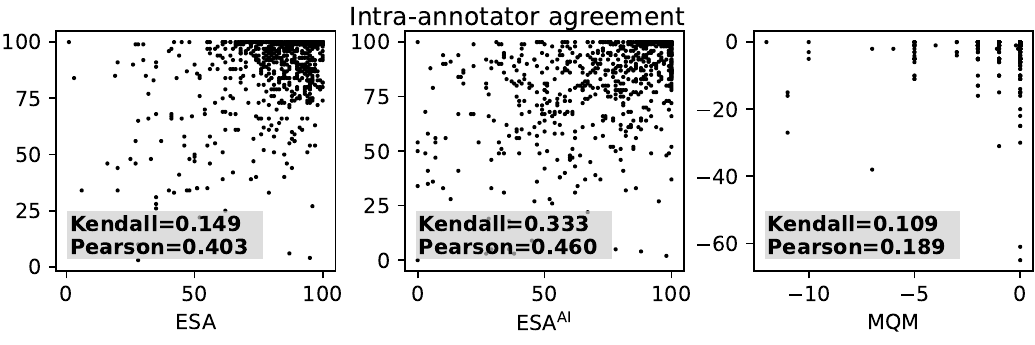}
\caption{Inter-annotator and intra-annotator agreement. The-intra annotator agreement shows changes in scoring by the same annotator when evaluated again. Each point represents single annotated segment with x-axis being annotator's score assigned one month and y-axis their score assigned two months later. For ESA and \esaai, the scores are directly from annotators. For MQM, they are computed by the formula. \esaai has the highest \textit{intra}-annotator and \textit{inter}-annotator agreement, showing another positive aspect of being primed by GEMBA.}
\label{fig:intrainter_annotator_agreement}
\end{figure*}

\begin{figure*}
\centering
\includegraphics[width=\linewidth]{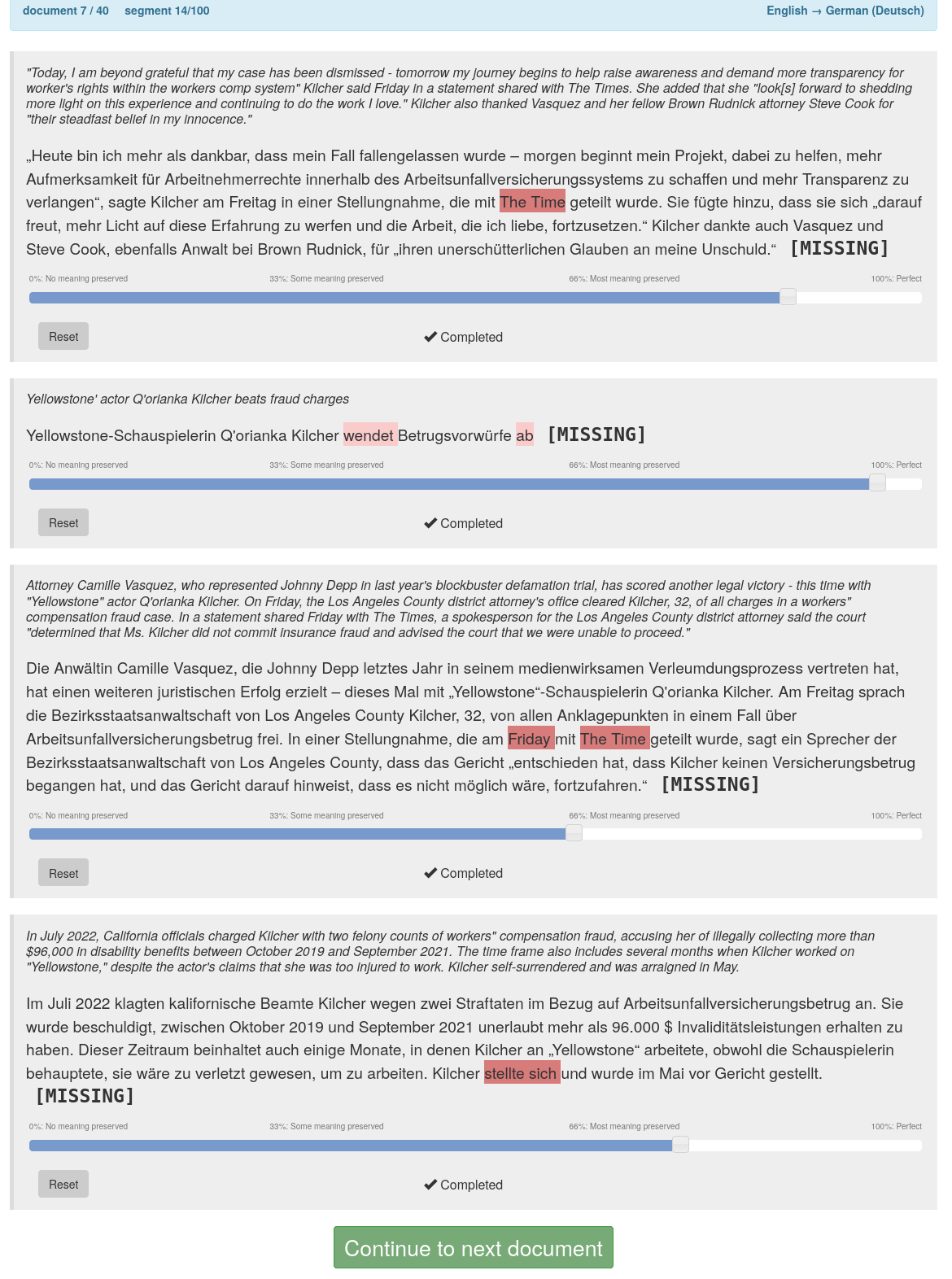}
\caption{Screenshot of the study interface implemented for Appraise. Multiple segments from a document are shown together for context. The AI suggests the initial error spans which the annotator post-edits and finally adds final score judgment.}
\label{fig:appraise_screenshot_full}
\end{figure*}

\begin{figure*}[htbp]
\section{GEMBA Quality Estimator Prompts}
\label{sec:gemba_prompts}

\begin{subfigure}[b]{\linewidth}
\begin{lstlisting}[breaklines,basicstyle=\ttfamily\fontsize{8.5}{9}\selectfont]
Your task is to identify machine translation errors and assess the tranlsation quality.

{source_lang} source:
```{source_seg}```
{target_lang} translation:
```{target_seg}```

Based on the source segment and machine translation surrounded with triple backticks,
identify error types in the translation and classify them. The categories of errors are:
accuracy (addition, mistranslation, omission, untranslated text), fluency (character enc-
oding, grammar, inconsistency, punctuation, register, spelling), style (awkward), termin-
ology (inappropriate for context, inconsistent use), non-translation, other, or no-error.

Each error is classified as one of two categories: major or minor. Major errors disrupt
the flow and make the understandability of text difficult or impossible. Minor errors
are errors that do not disrupt the flow significantly and what the text is trying to say
is still understandable.
\end{lstlisting}
\caption{Prompt for annotating error spans (initial step).}
\end{subfigure}
\begin{subfigure}[b]{\linewidth}
\begin{lstlisting}[breaklines,basicstyle=\ttfamily\fontsize{8.5}{9}\selectfont]
"source_lang": "English",
"source_seg": "I do apologise about this, we must gain permission from the account holder to discuss an order with another person, I apologise if this was done previously, however, I would not be able to discuss this with yourself without the account holders permission.",
"target_lang": "German",
"target_seg": "Ich entschuldige mich dafuer, wir muessen die Erlaubnis einholen, um eine Bestellung mit einer anderen Person zu besprechen. Ich entschuldige mich, falls dies zuvor geschehen waere, aber ohne die Erlaubnis des Kontoinhabers waere ich nicht in der Lage, dies mit dir involvement.",

"answer: """\
Major:
accuracy/mistranslation - "involvement"
accuracy/omission - "the account holder"
Minor:
fluency/grammar - "waere"
fluency/register - "dir"
"""
\end{lstlisting}
\caption{Prompt for scoring with prior annotations of error spans (example in prompt).}
\end{subfigure}
\begin{subfigure}[b]{\linewidth}
\begin{lstlisting}[breaklines,basicstyle=\ttfamily\fontsize{8.5}{9}\selectfont]
Given the translation from {source_lang} to {target_lang} and the annotated error
spans, assign a score on a continuous scale from 0 to 100. The scale has following
reference points: 0="No meaning preserved", 33="Some meaning preserved", 66="Most
meaning preserved and few grammar mistakes", up to 100="Perfect meaning and grammar".

Score the following translation from {source_lang} source:
```{source_seg}```
{target_lang} translation:
```{target_seg}```
Annotated error spans:
```{error_spans}```
Score (0-100): 
\end{lstlisting}
\caption{Prompt for scoring with prior annotations of error spans (final step).}
\end{subfigure}
\caption{Prompts for GEMBA with GPT-4. See the full \href{https://github.com/MicrosoftTranslator/GEMBA/blob/68be552640e2180bc0b6e1c3963592126a59b43c/gemba/gemba_esa.py}{GEMBA code for ESA}. The prompts can be used to first prompt GEMBA to produce the list of translation errors, as in MQM, and then prompt again to score the segments hollistically. For the \esaai human pre-annotations we only use the first part and only the error severities.}
\end{figure*}

\end{document}